\documentclass{article}

\PassOptionsToPackage{numbers, compress}{natbib}
\usepackage[dblblindworkshop,final]{neurips_2025}
\workshoptitle{The Second Workshop on GenAI for Health: Potential, Trust, and Policy Compliance}
\usepackage{amsmath}
\usepackage{graphicx}
\usepackage[utf8]{inputenc} 
\usepackage[T1]{fontenc}    
\usepackage{hyperref}       
\usepackage{url}            
\usepackage{booktabs}       
\usepackage{amsfonts}       
\usepackage{nicefrac}       
\usepackage{microtype}      
\usepackage{xcolor}         
\usepackage{wrapfig}
\usepackage{adjustbox}
 \usepackage{multirow}
\usepackage[table]{xcolor}
\usepackage{colortbl}
\usepackage{amsmath}
\usepackage{amssymb}
\usepackage{subcaption}  

\title{K-Stain: Keypoint-Driven Correspondence for H\&E-to-IHC Virtual Staining}

\author{
Sicheng Yang$^{1}$ \quad
Zhaohu Xing$^{1}$ \quad 
Haipeng Zhou$^{1}$ \quad
Lei Zhu$^{1,2}$\thanks{Lei Zhu (leizhu@ust.hk) is the corresponding author.} \\
$^1$The Hong Kong University of Science and Technology (Guangzhou)\\
$^2$The Hong Kong University of Science and Technology\\
}

\begin{document}

\maketitle

\begin{abstract}
Virtual staining offers a promising method for converting Hematoxylin and Eosin (H\&E) images into Immunohistochemical (IHC) images, eliminating the need for costly chemical processes. However, existing methods often struggle to utilize spatial information effectively due to misalignment in tissue slices. To overcome this challenge, we leverage keypoints as robust indicators of spatial correspondence, enabling more precise alignment and integration of structural details in synthesized IHC images. We introduce K-Stain, a novel framework that employs keypoint-based spatial and semantic relationships to enhance synthesized IHC image fidelity. K-Stain comprises three main components: (1) a Hierarchical Spatial Keypoint Detector (HSKD) for identifying keypoints in stain images, (2) a Keypoint-aware Enhancement Generator (KEG) that integrates these keypoints during image generation, and (3) a Keypoint Guided Discriminator (KGD) that improves the discriminator’s sensitivity to spatial details. Our approach leverages contextual information from adjacent slices, resulting in more accurate and visually consistent IHC images. Extensive experiments show that K-Stain outperforms state-of-the-art methods in quantitative metrics and visual quality. 
\end{abstract}

\section{Introduction}
Histopathological stain is an essential process in clinical
pathological analysis~\citep{boktor2022virtual}. Hematoxylin and Eosin (H\&E) and immunohistochemical (IHC) staining are two widely used staining methods. H\&E staining is a basic and cost-effective technique for observing tissue morphology and structure, but it lacks specificity for specific proteins~\citep{asaf2024dual}. IHC stain uses specific antibodies to identify target proteins in tissues, making it crucial for cancer diagnosis. Although IHC staining offers high specificity, it is more complex and expensive~\citep{anglade2020can}. These characteristics have motivated researchers to explore whether H\&E images can be computationally converted into IHC-like counterparts.

\begin{figure}[t]
\centering
\includegraphics[width=0.95\columnwidth]{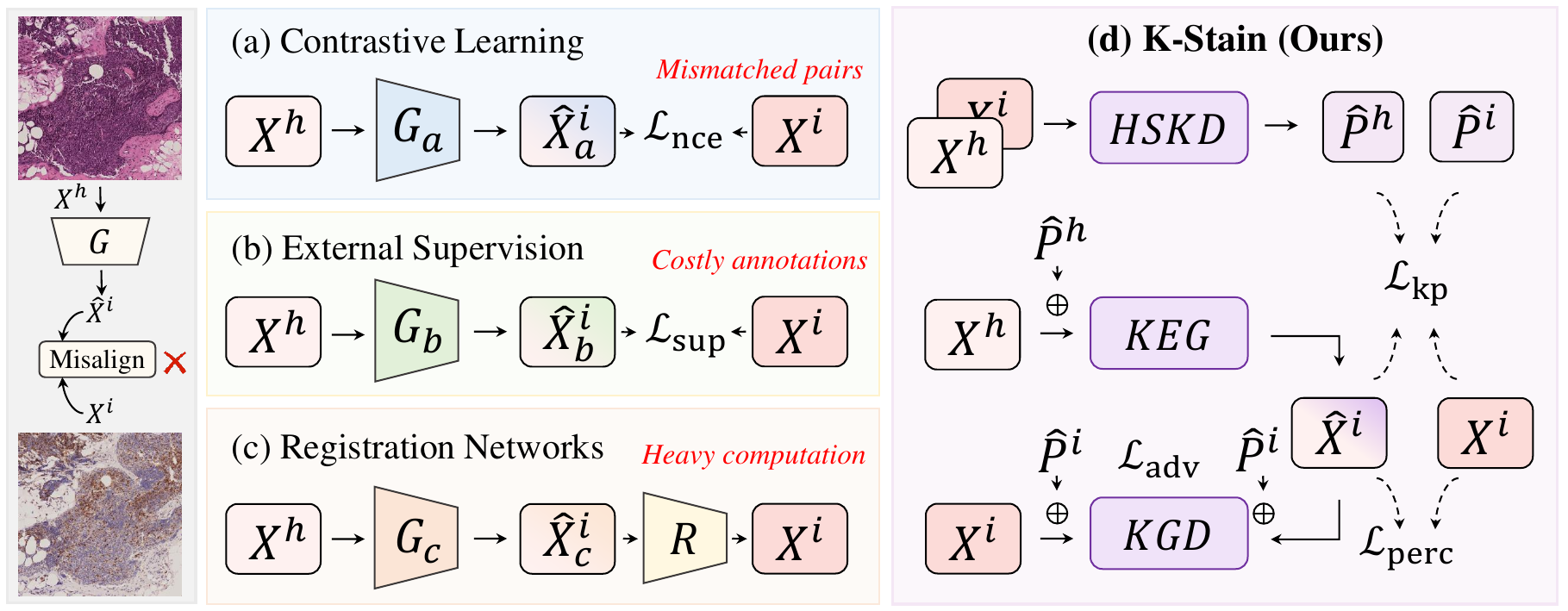}
\caption{Illustration of different strategies for handling misalignment in virtual staining. 
(a) Contrastive Learning, (b) External Supervision, (c) Registration Networks, and (d) our proposed K-Stain framework. 
Here, $X_h$ and $X_i$ denote the H\&E and IHC images, respectively; $\hat{X}_i$ is the generated IHC; 
$\hat{P}_h$ and $\hat{P}_i$ are the predicted keypoints; $\mathcal{L}_{\text{kp}}$, $\mathcal{L}_{\text{perc}}$, and $\mathcal{L}_{\text{adv}}$ 
represent the keypoint-guided reconstruction, perceptual, and adversarial losses, respectively.
}
\label{fig:intro}
\vspace{-2mm}
\end{figure}

Digital pathology is rapidly evolving, and numerous researchers are developing various virtual staining methods based on deep learning~\citep{xing2022nestedformer,xing2024hybrid,xing2025detect,xing2024cross,xing2025segmamba,xing2024segmamba,wang2024video,wang2024dual,wang2025serp,wang2025versatile,wang2025toward}. These methods enable the generation of one type of stain from another without the need for chemical staining procedures. However, many traditional generative models cannot adapt to a key characteristic of virtual staining. In practice, the paired images are obtained from two consecutive tissue sections that are stained separately. Although these sections are adjacent, they are not identical since cells and structures may appear in one section but not in the other, and their spatial arrangement inevitably shifts due to tissue slicing. Consequently, the dataset lacks truly aligned image pairs, which makes pixel-wise supervision (e.g., $\ell_{1}$ or $\ell_{2}$ loss) unreliable. This intrinsic misalignment poses a fundamental challenge for conventional generative frameworks that rely on strictly paired data~\citep{li2023adaptive}.
To address this issue, there are generally three approaches: 
(a) Contrastive Learning, which constructs corresponding patch pairs to align semantically similar patches~\citep{li2023adaptive,chen2024pathological,li2024exploiting,zhang2024high} (Fig.~\ref{fig:intro}a). 
(b) External Supervision Signals, such as segmentation maps~\citep{hu2024boosting}, nuclei density graph~\citep{chen2024pathological} or protein prototype~\citep{peng2024advancing} (Fig.~\ref{fig:intro}b). 
(c) An additional registration network, which calibrates the spatial position of generated images~\citep{li2024virtual} (Fig.~\ref{fig:intro}c). 

 However, contrastive learning methods only enforce semantic alignment, which prevents them from fully exploiting the fine-grained spatial correspondence available in paired data. This often results in mismatches for subtle but clinically important structures, such as nuclei boundaries or glandular morphology. External supervision signals can provide useful auxiliary constraints, yet they do not explicitly address spatial misalignment and their acquisition, such as segmentation maps, nuclei density graphs, or protein prototypes, is labor-intensive and costly, limiting practical scalability. Registration networks aim to resolve alignment by calibrating spatial positions, but they introduce considerable computational burden, particularly when processing high-resolution whole-slide images.

To address these limitations, we introduce \textbf{K-Stain}, a novel framework that explicitly integrates keypoint-based spatial cues into the virtual staining process (Fig.~\ref{fig:intro}d). Keypoints act as compact and discriminative descriptors of structural landmarks, effectively capturing local correspondences between different stains. Leveraging these spatial anchors enables the network to align subtle tissue structures more accurately without relying on dense annotations or computationally expensive registration. By embedding keypoint relationships into the generation pipeline, K-Stain achieves improved spatial consistency and reduced computational cost.

Specifically, K-Stain consists of three components. First, we introduce a Hierarchical Spatial Keypoint Detector (HSKD) for efficient keypoint prediction in HE and IHC images, capturing both spatial and semantic relationships. Second, we propose a Keypoint-aware Enhancement Generator (KEG) that leverages keypoint information to guide the generation process, improving spatial consistency and enhancing the quality of synthesized IHC images. Third, we design a Keypoint Guided Discriminator (KGD) that incorporates keypoints to distinguish real from synthetic images, enabling the discriminator to focus simultaneously on semantic content and spatial details. Extensive experiments on two datasets demonstrate that K-Stain significantly enhances virtual staining quality and achieves superior performance compared with state-of-the-art methods.

\section{Related Work}

\paragraph{Deep learning for virtual staining.}
A variety of deep learning approaches have been developed to enable virtual staining between H\&E and IHC images. Early models, such as CycleGAN-like frameworks~\citep{shaban2019stainGAN,zhu2017unpaired,wang2024cytogan} and Pix2Pix-like models~\citep{wang2018high,vasiljevic2023histostargan}, demonstrated the feasibility of stain conversion, yet their dependence on pixel-level supervision made them susceptible to slice misalignment. To alleviate this issue, contrastive learning objectives were introduced to enhance semantic alignment across patches~\citep{li2023adaptive,zhang2024high}; however, inaccurate correspondences often led to the loss of clinically relevant fine structures. To preserve structural fidelity, Dubey \textit{et al.}~\citep{dubey2023structural} proposed SC-GAN, which integrates edge-based priors, decoder-side attention, and a structural loss to maintain contextual integrity, though its reliance on handcrafted features restricts generalization. Zeng \textit{et al.}~\citep{zeng2022semi} further developed a semi-supervised PR-staining method that combines patch-level labels derived from registration with a classifier to enforce pathological consistency, but the requirement for precise registration remains a key limitation. More recently, diffusion-based frameworks have emerged as a promising alternative. Shen \textit{et al.}~\citep{shen2023staindiff} introduced StainDiff, a probabilistic diffusion framework for virtual staining, while Jewsbury \textit{et al.}~\citep{jewsbury2024stainfuser} proposed StainFuser, a conditional latent diffusion model trained on large-scale pathology data.

Efficiently capturing fine-grained structural correspondence under misaligned conditions remains an open challenge.

\paragraph{Keypoint detection in medical imaging.}
Keypoint representations provide a compact means of encoding anatomical landmarks and establishing spatial correspondence, and have been widely adopted in medical image analysis. In segmentation tasks~\citep{yi2019multi,yang2024keypoint}, keypoints serve as localized anchors that facilitate instance separation by grouping predefined points into structured objects, while also enriching features with long-range dependencies through keypoint-augmented fusion layers. In registration tasks~\citep{hansen2021graphregnet,wang2023robust}, sparse yet distinctive keypoints act as stable geometric constraints, enabling accurate estimation of dense deformation fields while alleviating the memory burden by obviating the need for additional registration networks. In the context of virtual staining, adjacent tissue slices exhibit strong local similarity, and paired H\&E and IHC images naturally contain matching points that preserve structural semantics. Although slight pixel shifts or deformations may occur between slices, their glandular layout and cellular organization remain largely consistent within local regions. K-Stain leverages this property by detecting and matching keypoints to capture robust structural correspondences under small misalignments.

\section{Methodology}

\subsection{Overview}
Our proposed framework, K-Stain, addresses the misalignment challenge in virtual staining by incorporating spatial correspondence through keypoints. As illustrated in Fig.~\ref{fig:framework}, K-Stain consists of three main modules: (1) a Hierarchical Spatial Keypoint Detector (HSKD) that adaptively learns to predict spatially and semantically consistent keypoints between H\&E and IHC images in an end-to-end manner, (2) a Keypoint-aware Enhancement Generator (KEG) that embeds the detected keypoints into dense feature maps to guide the generation process, thereby enhancing structural fidelity, and (3) a Keypoint Guided Discriminator (KGD) that leverages keypoint-derived structural priors to enforce consistent adversarial supervision. These components are jointly optimized under a combination of keypoint-guided reconstruction loss, perceptual loss, and adversarial loss, which together enforce spatial alignment, semantic consistency, and realistic appearance in the synthesized IHC images. This adaptive and modular design enables K-Stain to effectively enhance both the quality and structural consistency of virtual staining compared with conventional generative approaches.

\subsection{Hierarchical Spatial Keypoint Detector}
\begin{figure}[t]
\centering
\includegraphics[width=0.95\columnwidth]{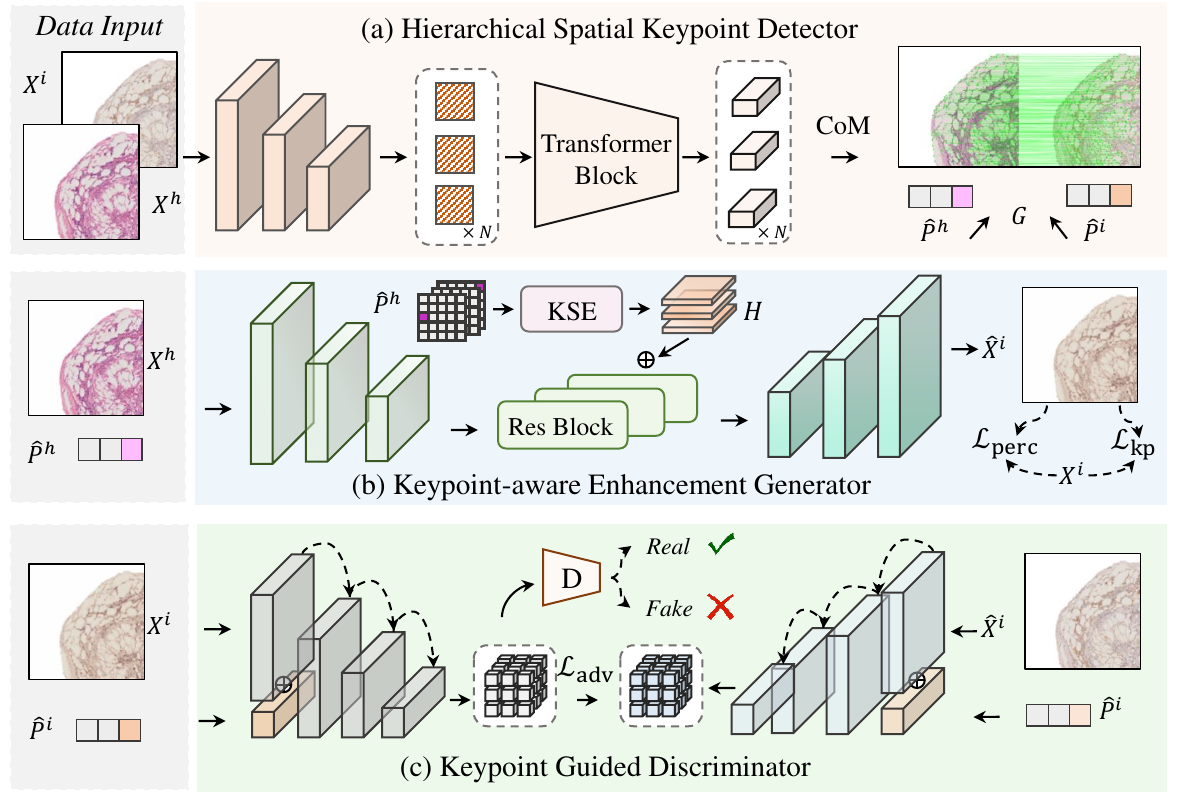}
\caption{
The proposed K-Stain framework integrates keypoint-based spatial correspondence to address misalignment in virtual staining. It consists of (a) a Hierarchical Spatial Keypoint Detector (HSKD) that predicts consistent keypoints and estimates affine transformation, (b) a Keypoint-aware Enhancement Generator (KEG) that embeds keypoints into dense feature maps for IHC synthesis, and (c) a Keypoint Guided Discriminator (KGD) that enforces consistent adversarial supervision. 
}
\label{fig:framework}
\end{figure}

We introduce the Hierarchical Spatial Keypoint Detector (HSKD) to predict keypoints from paired H\&E and IHC images, as shown in Fig.~\ref{fig:framework}(a). Given the input images $X^{h}$ and $X^{i}$, we first process them with a series of Convolutional Neural Network (CNN) layers to extract hierarchical feature representations. These feature maps are subsequently embedded into a sequence of tokens, denoted as $\{z_{1}, z_{2}, \dots, z_{N}\}$, where $N$ is the number of tokens.

To enhance keypoint localization by capturing long-range dependencies, the token sequence is processed by a Transformer block~\cite{vaswani2017attention}. Specifically, the multi-head self-attention (MHSA) mechanism computes pairwise interactions among tokens. Given the query, key, and value projections:
\begin{align}
Q = ZW^{Q}, \quad K = ZW^{K}, \quad V = ZW^{V}, \\
\text{Attention}(Q, K, V) = \text{Softmax}\left(\frac{QK^{\top}}{\sqrt{d}}\right)V,
\end{align}
where \(Z \in \mathbb{R}^{N \times d}\) represents the token embeddings, \(W^{Q}, W^{K}, W^{V} \in \mathbb{R}^{d \times d}\) are learnable projection matrices, and \(d\) is the embedding dimension. The attention-enhanced features are then updated as
\begin{align}
Z' = \text{MHSA}(Z).
\end{align}

Finally, the enhanced feature maps are passed through a Center of Mass (CoM) operator~\cite{ma2020volumetric}, which converts the feature activations into spatial coordinates. This yields two sets of predicted keypoints for the HE and IHC images, denoted as $\hat{P}^{h}, \hat{P}^{i} \in \mathbb{R}^{N \times 2}$, where each row corresponds to the \((x,y)\)-coordinate of a keypoint.

Based on the predicted keypoints $\hat{P}^{h}$ and $\hat{P}^{i}$, 
we directly estimate an affine transformation that aligns $X^{h}$ to $X^{i}$. 
The optimal affine matrix $A$ is obtained via a least-squares fit:
\begin{align}
A = \arg\min_{M} 
\big\| M \tilde{P}^{h} - \tilde{P}^{i} \big\|_{F}^{2},
\end{align}
where $\tilde{P}^{h}$ and $\tilde{P}^{i}$ denote the homogeneous coordinates of $\hat{P}^{h}$ and $\hat{P}^{i}$, and $\|\cdot\|_{F}$ is the Frobenius norm. 
To warp the HE image, we construct a differentiable sampling grid $\mathcal{G}$ in normalized coordinates using the inverse affine matrix $A^{-1}$:
\begin{align}
\mathcal{G}(p) = A^{-1}p,
\end{align}
where $p$ denotes a pixel coordinate in the target (IHC) space. 
The warped HE image is then obtained by bilinear sampling:
\begin{align}
X^{h}_{\text{w}}(p) = X^{h}\big(\mathcal{G}(p)\big).
\end{align}

The HSKD module jointly learns keypoint consistency and affine transformation, suppressing unreliable correspondences by optimizing spatially consistent keypoint pairs instead of raw pixels. 
Being fully differentiable, it enables adaptive keypoint learning without requiring a separate registration network. 
The resulting keypoints and alignment matrix are subsequently utilized by KEG (Sec.~\ref{sec:KEG}) and KGD (Sec.~\ref{sec:KGD}) for generation and discrimination.

\subsection{Keypoint-aware Enhancement Generator}
\label{sec:KEG}
We propose the Keypoint-aware Enhancement Generator (KEG), which leverages spatial information from predicted keypoints to guide the synthesis of IHC images. 
As shown in Fig.~\ref{fig:framework}(b), the detected keypoints $\hat{P}^{h}$ are first converted into the image domain by the Keypoint Spatial Embedding (KSE) module (see Fig.~\ref{fig:kse}). Instead of performing registration directly in the feature space, we project keypoints back into the image domain through the KSE module. 
This mapping preserves spatial interpretability and avoids high-dimensional feature-space registration, which would otherwise incur significant computational overhead and hinder end-to-end optimization. KSE projects each $\hat{p}_i \in \hat{P}^{h}$ onto a 2D grid using a localized kernel $\phi(\cdot)$.
In our experiments, we adopt a Gaussian kernel:
\begin{align}
h_i(x, y) = \phi \big((x, y), \hat{p}_i; \sigma\big),
\end{align}
where $\sigma$ controls the spatial influence. 
The Gaussian kernel is chosen because it provides smooth and spatially localized activations, effectively modeling the gradual structural variations in tissue and avoiding sharp discontinuities that may arise from alternative kernels~\citep{chung2020gaussian}. 
Each heatmap $h_i$ is then passed through a $1\times1$ convolution to obtain $\tilde{h}_i$, and all embeddings are concatenated:
\begin{align}
H = \mathrm{Concat}(\tilde{h}_1, \tilde{h}_2, \ldots, \tilde{h}_N),
\end{align}
producing a dense tensor $H$ that encodes the structural layout indicated by keypoints.

In parallel, the H\&E image $X^{h}$ is encoded into multi-scale feature maps through a downsampling path:
\begin{align}
F_{\text{down}} = \mathrm{Down}(X^{h}),
\end{align}
which reduces spatial resolution while preserving semantic context. 
The ResBlock pathway then integrates structural priors by concatenating $F_{\text{down}}$ with $H$:
\begin{align}
F_{\text{res}} = \mathrm{ResBlock}\big(\mathrm{Concat}(F_{\text{down}}, H)\big),
\end{align}
where the residual blocks~\citep{he2016deep} refine the fused representation by maintaining stable gradient flow and enhancing feature discrimination. 
This design allows the model to capture both appearance information and keypoint-guided structural information. 
Finally, an upsampling path reconstructs the virtual IHC image:
\begin{align}
\hat{X}^{i} = \mathrm{Up}(F_{\text{res}}).
\end{align}

By embedding sparse keypoints into dense feature maps and integrating them with residual appearance features, KEG leverages the spatial correspondence encoded by keypoints to ensure alignment and improve the fidelity of virtual staining.

\subsection{Keypoint Guided Discriminator}
\label{sec:KGD}
The Keypoint Guided Discriminator (KGD) integrates spatial keypoint information into the adversarial training process. 
Specifically, the keypoint set is first embedded into dense heatmaps through the Keypoint-to-Structural Embedding (KSE) module, yielding a structural tensor $H$. 
As shown in Fig.~\ref{fig:framework}(c), the discriminator then receives both the IHC image (\(X^{i}\) or \(\hat{X}^{i}\)) and the corresponding structural tensor $H$. 
These tensors are fused with image features at multiple scales within the discriminator backbone, ensuring that structural correspondence is explicitly encoded during discrimination. Formally, the discriminator is defined as
\begin{align}
D(X, H) \to [0,1],
\end{align}
where \(X\) denotes either a real IHC image \(X^{i}\) or a generated image \(\hat{X}^{i}\), and \(H=\mathrm{KSE}(P)\) represents the structural embedding derived from its associated keypoints. 

We adopt a conditional non-saturating GAN objective~\citep{goodfellow2014generative}. The discriminator aims to assign high scores to real pairs and low scores to generated pairs:
\begin{align}
\mathcal{L}_{D} &= -\,\mathbb{E}_{(X^{i},P^{i})}\!\left[\log D\!\big(X^{i}, H\big)\right]
                  -\,\mathbb{E}_{(\hat{X}^{i},\hat{P}^{i})}\!\left[\log\!\big(1 - D\big(\hat{X}^{i}, H\big)\big)\right], \\
\mathcal{L}_{G} &= -\,\mathbb{E}_{(\hat{X}^{i},\hat{P}^{i})}\!\left[\log D\big(\hat{X}^{i}, H\big)\right],
\end{align}
 Here, the generator \(G\) corresponds to the Keypoint-aware Enhancement Generator (KEG) introduced in Sec.~\ref{sec:KEG}. By introducing keypoint information, the discriminator evaluates not only global appearance realism but also the consistency of local structures around keypoint-defined regions. 
This design enforces that generated IHC images \(\hat{X}^{i}\) preserve semantic and spatial information represented by keypoints, thereby providing more reliable and structure-aware adversarial feedback.

\subsection{Optimization Objective}
The training objective of our framework combines keypoint-guided reconstruction loss, perceptual loss, and adversarial loss. 
These complementary objectives jointly enforce spatial alignment, semantic consistency, and visual realism in the generated IHC images.

\paragraph{Keypoint-guided reconstruction loss.}
To explicitly enforce spatial correspondence, we employ the affine transformation estimated in the HSKD module. 
Based on the predicted keypoints, HSKD yields an affine matrix $A$, which is used to warp the H\&E image into the IHC coordinate space, resulting in $X^{h}_{\text{w}}$. 
The warped image is then passed through the generator $G(\cdot)$, i.e., the Keypoint-aware Enhancement Generator (KEG), to synthesize the virtual IHC image. 
We compute an $\ell_{1}$ loss between the generated image and the ground truth IHC:
\begin{align}
\mathcal{L}_{\text{kp}} = \big\| G\!\big(X^{h}_{\text{w}}\big) - X^{i} \big\|_{1}.
\end{align}
Compared with a direct pixel-wise $\ell_{1}$ loss, $\mathcal{L}_{\text{kp}}$ explicitly leverages the spatial correspondence established by HSKD, thereby reducing misalignment artifacts and improving semantic consistency.

\paragraph{Perceptual loss.}
To further enhance visual quality, we employ a perceptual loss $\varphi_{\text{perc}}$ based on a pretrained feature extractor~\citep{johnson2016perceptual}. 
Specifically, we adopt the widely used VGG16 network~\citep{simonyan2014very} to compute high-level feature representations. 
The perceptual loss is defined as the $\ell_{2}$ distance between the features of the generated IHC image and those of the ground truth:
\begin{align}
\mathcal{L}_{\text{perc}} = \big\| \varphi(\hat{X}^{i}) - \varphi(X^{i}) \big\|_{2}^{2},
\end{align}
where $\varphi(\cdot)$ denotes the feature representation extracted by the fixed VGG16 network~\citep{simonyan2014very}. 
This encourages the generator to preserve semantic structures and stain-specific texture details that may not be captured by pixel-level supervision alone.
\begin{figure}[t]
    \centering
    \begin{minipage}{0.375\textwidth}
        \centering
        \includegraphics[width=\linewidth]{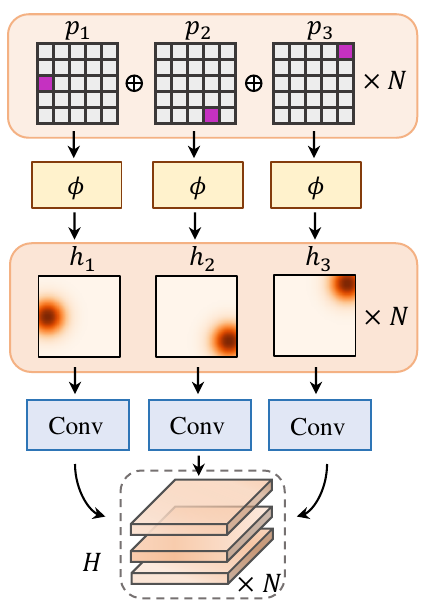}
        \caption{Illustration of the Keypoint Spatial Embedding (KSE) module, which encodes keypoint information into feature maps for virtual staining.}
        \label{fig:kse}
    \end{minipage}%
    \hfill
    \begin{minipage}{0.5\textwidth}
        \centering
        \includegraphics[width=\linewidth]{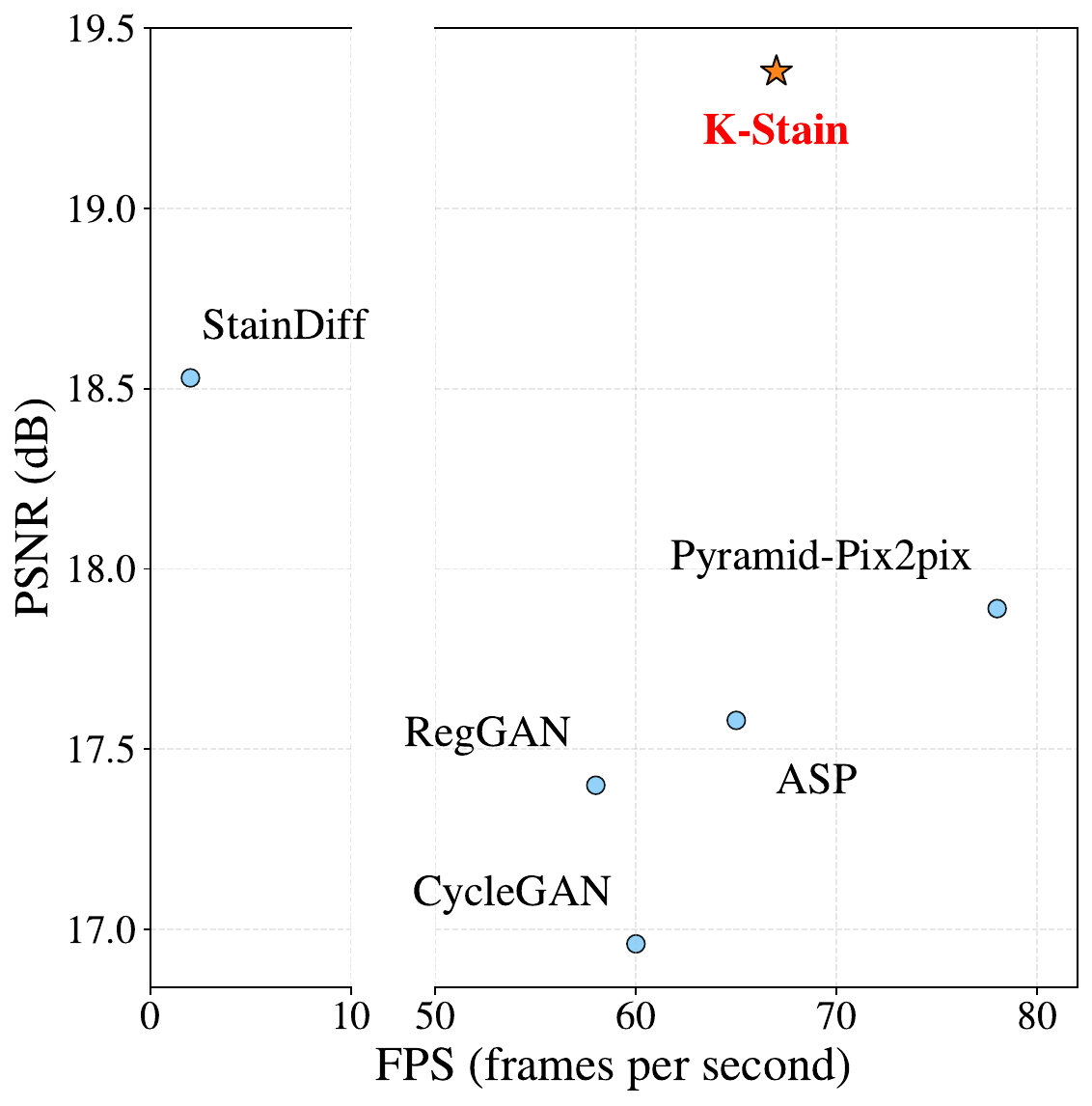}
        \caption{Comparison of model performance in terms of PSNR versus inference speed (FPS). The proposed K-Stain achieves the best trade-off between performance and efficiency compared to GAN- and diffusion-based baselines.}

        \label{fig:psnr_fps}
    \end{minipage}
\end{figure}

\paragraph{Adversarial loss.}
Finally, we introduce an adversarial loss~\citep{goodfellow2014generative} with the Keypoint Guided Discriminator (KGD). The discriminator is conditioned on both the synthesized IHC image and the structural tensor derived from the detected keypoints. By jointly evaluating the image appearance and the encoded spatial information, KGD enforces consistency between visual realism and the spatial relationships. This adversarial training objective drives the generator to produce IHC images that are not only indistinguishable from real ones, but also spatially aligned with the tissue morphology.

\paragraph{Overall objective.}
The overall generator loss is formulated as
\begin{align}
\mathcal{L} = \mathcal{L}_{\text{adv}} + \lambda_{\text{perc}}\mathcal{L}_{\text{perc}} + \lambda_{\text{kp}}\mathcal{L}_{\text{kp}},
\end{align}
where $\lambda_{\text{perc}}$ and $\lambda_{\text{kp}}$ are trade-off hyperparameters, set to 1 and 5, respectively, as these values yield the best performance in our ablation study (Sec.~\ref{sec:ab_study}).

\section{Experiments}
\subsection{Datasets and Implementation Details}
\paragraph{Datasets.} 
We use two publicly available datasets to evaluate K-Stain, namely the Breast Cancer Immunohistochemical (BCI) dataset~\cite{liu2022bci} and the H\&E to IHC image Translation (HIT) dataset~\cite{zhang2024high}. 
The BCI dataset has 3,896 image pairs for training and 977 image pairs for testing. 
For the HIT dataset, we focus on its CD3 section, which includes 1,652 image pairs for training and 155 image pairs for testing. 
We follow the official train-test split provided by the respective datasets.

\paragraph{Implementation Details.} 
\label{sec:details}
To ensure fair comparisons, we keep the discriminator architecture consistent with CycleGAN~\cite{zhu2017unpaired}. Specifically, we utilize 6 residual blocks in the generator. Our model is implemented in PyTorch 2.1.0 with CUDA 12.1 support. 
For data preprocessing, we perform random cropping to a resolution of 512×512. 
The batch size is set to 4 per GPU for each dataset. 
All experiments are conducted on a single NVIDIA A6000 GPU. 
We employ the Adam optimizer with an initial learning rate of $2\times10^{-4}$ and train the network for 100 epochs.

\paragraph{Evaluation Metrics.} 
We adopt the Peak Signal-to-Noise Ratio (PSNR), the Structural Similarity Index (SSIM), and the Learned Perceptual Image Patch Similarity (LPIPS)~\cite{zhang2018unreasonable} to quantitatively evaluate different methods. 
In addition to reconstruction quality, we also measure the inference efficiency of each method using Frames Per Second (FPS), which reflects the number of images the model can process per second during inference. 

\begin{table*}[!t]
    \centering
    \caption{Quantitative comparison. The \textbf{best} and the \underline{second-best} results are highlighted.}
    \label{tab:comp_sota}
    \renewcommand\arraystretch{1.4}
    \resizebox{0.98\textwidth}{!}{
    \begin{tabular}{c | c c c | c c c | c c c}
    \toprule
    \multirow{2}{*}{Methods} & \multicolumn{3}{c|}{BCI} & \multicolumn{3}{c|}{HIT}  & \multicolumn{3}{c}{Avg} \\
    & SSIM $\uparrow$ & PSNR $\uparrow$ & LPIPS $\downarrow$ & SSIM $\uparrow$ & PSNR $\uparrow$ & LPIPS $\downarrow$ & SSIM $\uparrow$ & PSNR $\uparrow$ & LPIPS $\downarrow$\\
    \midrule

CycleGAN
& 0.4739 & 16.68 & 0.2918 & 0.3838 & 17.24 & 0.2437 & 0.4289 & 16.96 & 0.2678 \\
Pix2pix
& 0.4636 & 16.64 & 0.2947 & 0.3594 & 16.96 & 0.2478 & 0.4115 & 16.80 & 0.2713 \\
Pix2pixHD
& 0.4685 & 16.84 & 0.2934 & 0.3622 & 17.47 & 0.2337 & 0.4153 & 17.16 & 0.2636 \\
GcGAN
& 0.4720 & 16.98 & 0.2924 & 0.3922 & 17.90 & 0.2373 & 0.4327 & 17.44 & 0.2648 \\
CUT
& 0.4752 & 17.05 & 0.2911 & 0.3594 & 17.37 & 0.2468 & 0.4173 & 17.21 & 0.2690 \\
RegGAN
& 0.4701 & 16.99 & 0.2920 & 0.3720 & 17.82 & 0.2351 & 0.4210 & 17.40 & 0.2635 \\
Pyramid-Pix2pix
& 0.4881 & 17.90 & 0.2977 & 0.3933 & 17.89 & 0.2373 & 0.4407 & 17.89 & 0.2675 \\
ASP
& 0.4992 & 17.50 & 0.2949 & \underline{0.3963} & 17.66 & 0.2463 & 0.4478 & 17.58 & 0.2706 \\
StainFuser
& 0.5072 & 18.12 & \underline{0.2750} & 0.3904 & 17.62 & 0.2486 & 0.4487 & 17.86 & 0.2538 \\
StainDiff
& \underline{0.5113} & \underline{18.43} & 0.2809 & 0.3932 & \underline{18.66} & \underline{0.2142} & \underline{0.4515} & \underline{18.53} & \underline{0.2471} \\
\midrule

\rowcolor{gray!20} \textbf{K-Stain} 
& \textbf{0.5268} & \textbf{19.82} & \textbf{0.2665} 
& \textbf{0.4162} & \textbf{18.93} & \textbf{0.2061} 
& \textbf{0.4720} & \textbf{19.38} & \textbf{0.2361} \\

\bottomrule
\end{tabular}
}
\end{table*}

\subsection{Comparison with SOTA Methods}
We compare the proposed method with two representative categories of virtual staining approaches: (a) GAN-based methods, including CycleGAN~\cite{zhu2017unpaired}, Pix2pix~\cite{isola2017image}, Pix2pixHD~\cite{wang2018high}, GcGAN~\cite{fu2019geometry}, CUT~\cite{park2020contrastive}, RegGAN~\cite{kong2021breaking}, Pyramid-Pix2pix~\cite{liu2022bci}, and ASP~\cite{li2023adaptive}; 
and (b) diffusion-based methods, represented by StainFuser~\cite{jewsbury2024stainfuser} and StainDiff~\cite{shen2023staindiff}.

\paragraph{Quantitative Comparisons.} 
Table~\ref{tab:comp_sota} reports the quantitative results of different methods on the BCI and HIT datasets in terms of SSIM, PSNR, and LPIPS. Compared with all competing approaches, our proposed K-Stain consistently achieves the best performance across both datasets. On BCI, K-Stain improves SSIM and PSNR by +1.55\% and +1.39 dB over the second-best method, while also reducing LPIPS by --0.0085. On HIT, K-Stain attains clear gains as well, outperforming the next best competitor in SSIM by +1.99\% and in PSNR by +0.27 dB, along with the lowest LPIPS (0.2061). Averaged across datasets, K-Stain reaches 0.4720 in SSIM, 19.38 dB in PSNR, and 0.2361 in LPIPS, surpassing the second-best results  by +2.05\%, +0.85 dB, and --0.011, respectively. These results highlight the superiority of K-Stain over existing virtual staining methods.
\begin{figure}[t]
\centering
\includegraphics[width=0.95\columnwidth]{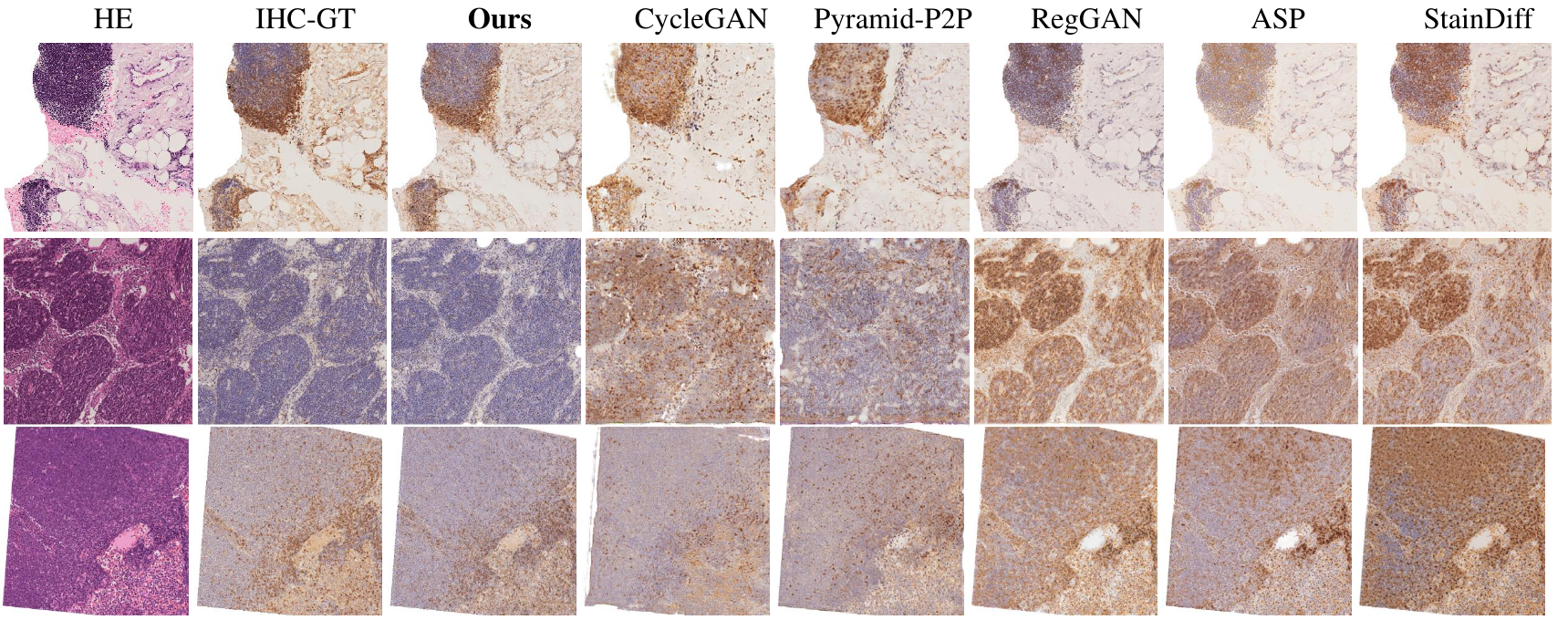}
\caption{
 Visual comparisons of proposed K-Stain and other methods.
} 
\label{fig:vis}
\end{figure}
\paragraph{Visual Comparisons.}
Moreover, we provide visual comparisons between K-Stain and state-of-the-art methods on the BCI and HIT datasets. 
As shown in Figure~\ref{fig:vis}, our approach produces synthesized IHC images that better preserve fine-grained cellular morphology and large-scale tissue architecture, avoiding the structural distortions or artifacts often observed in other methods. 
In addition, the staining appearance generated by our framework is more consistent with the ground truth, exhibiting natural color distribution and clearer boundary delineation. 
These visual comparisons further demonstrate the effectiveness of our model in achieving both structural fidelity and staining realism.

\paragraph{Efficiency Comparisons.} 
As shown in Fig.~\ref{fig:psnr_fps}, the proposed K-Stain achieves the most favorable balance between accuracy and inference efficiency compared with GAN- and diffusion-based baselines. Specifically, K-Stain attains the highest PSNR of 19.38~dB while sustaining a fast inference speed of 67 FPS. In contrast, diffusion-based StainDiff achieves a relatively high PSNR of 18.53~dB but suffers from extremely low efficiency (2 FPS). GAN-based approaches such as CycleGAN, RegGAN, ASP, and Pyramid-Pix2pix provide higher inference speeds (58--78 FPS) but remain clearly inferior in terms of PSNR (16.96--17.89~dB). These results highlight that K-Stain consistently achieves the best trade-off between accuracy and efficiency
, outperforming both GAN and diffusion counterparts.

\subsection{Ablation Study}
\label{sec:ab_study}
\begin{table}[t]
\centering
\caption{Ablation study for different modules on
BCI and HIT dataset.}
\small
\begin{tabular}{l|cc|ccc}
\toprule
Methods & KEG& KGD& SSIM $\uparrow$ & PSNR $\uparrow$ & LPIPS $\downarrow$ \\
\midrule
Basic &  & & 0.4289	& 16.96	& 0.2678 \\
M1 &  \checkmark& & 0.4465 & 18.52 & 0.2470 \\
M2 &  &\checkmark & 0.4360 & 17.80 & 0.2535 \\
Ours w/o HSKD & \checkmark & \checkmark & 0.4380 & 17.95 & 0.2620 \\
Ours &  \checkmark & \checkmark & \textbf{0.4720} & \textbf{19.38} & \textbf{0.2361}\\
\bottomrule
\end{tabular}
\label{tab:ablation}
\end{table}

\begin{table*}[t]
\centering
\begin{minipage}{0.48\textwidth}
\centering
\caption{Ablation study on the trade-off hyperparameters 
$\lambda_{\text{perc}}$ and $\lambda_{\text{kp}}$ in the loss function.}
\small
\begin{tabular}{c c | c c c}
\toprule
$\lambda_{\text{perc}}$ & $\lambda_{\text{kp}}$ & SSIM $\uparrow$ & PSNR $\uparrow$ & LPIPS $\downarrow$ \\
\midrule
1   & 1   & 0.4685 & 19.15 & 0.2395 \\
1   & 5   & \textbf{0.4720} & \textbf{19.38} & \textbf{0.2361} \\
1   & 10  & 0.4698 & 19.20 & 0.2384 \\
5   & 1   & 0.4662 & 19.02 & 0.2412 \\
10  & 1   & 0.4650 & 18.95 & 0.2420 \\
\bottomrule
\end{tabular}
\label{tab:lambda_ablation}
\end{minipage}%
\hfill
\begin{minipage}{0.48\textwidth}
\centering
\caption{Ablation study on the Gaussian kernel parameter $\sigma$ (smoothing strength).}
\small
\begin{tabular}{c|c c c}
\toprule
$\sigma$ & SSIM $\uparrow$ & PSNR $\uparrow$ & LPIPS $\downarrow$ \\
\midrule
0.1  & 0.4655 & 18.95 & 0.2432 \\
0.5  & 0.4695 & 19.20 & 0.2385 \\
1.0  & \textbf{0.4720} & \textbf{19.38} & \textbf{0.2361} \\
5.0  & 0.4678 & 19.02 & 0.2403 \\
10.0 & 0.4542 & 17.90 & 0.2840 \\
\bottomrule
\end{tabular}
\label{tab:sigma_ablation}
\end{minipage}
\end{table*}

\paragraph{Module Ablation.} 
We conduct an ablation study to assess the contributions of HSKD, KEG, and KGD, 
where the baseline (``Basic'') is a GAN~\cite{goodfellow2014generative} implemented with residual convolutional blocks. As summarized in Table~\ref{tab:ablation}, introducing KEG into the baseline (``M1'') increases SSIM by approximately +2\% and PSNR by +1.56 dB, accompanied by a notable reduction in LPIPS. Similarly, substituting the baseline discriminator with KGD (``M2'') brings a +0.7\% improvement in SSIM and +1.11 dB in PSNR, while also slightly lowering LPIPS. When both KEG and KGD are jointly integrated, the model achieves the most significant gains. In contrast, removing the HSKD module (``Ours w/o HSKD'') results in a substantial degradation: SSIM decreases by --3.4\%, PSNR drops by --1.43 dB, and LPIPS increases by +0.026 relative to the full model. These findings collectively demonstrate that each module contributes positively, while their combination produces the most pronounced performance improvements
(see Appendix~\ref{appendix:appB} for additional ablation results).

\paragraph{Hyperparameter Ablation.} 
We conduct ablation experiments to investigate how different hyperparameter settings influence the performance of K-Stain. As shown in Table~\ref{tab:lambda_ablation}, balancing the perceptual and keypoint losses is essential. Setting $\lambda_{\text{perc}}=1$ and $\lambda_{\text{kp}}=5$ yields the best results, improving SSIM by +0.35\% and PSNR by +0.23 dB compared to the equal-weight case ($\lambda_{\text{perc}}=\lambda_{\text{kp}}=1$), while also reducing LPIPS. Excessively increasing $\lambda_{\text{kp}}$ to 10 or $\lambda_{\text{perc}}$ to 5 or 10 leads to consistent performance drops, indicating that an imbalanced loss trade-off can weaken structural guidance. Regarding the Gaussian kernel parameter (Table~\ref{tab:sigma_ablation}), a moderate smoothing strength ($\sigma=1.0$) achieves the best overall results. Smaller values (e.g., $\sigma=0.1, 0.5$) slightly reduce structural fidelity, while overly large values (e.g., $\sigma=10.0$) cause a pronounced degradation, with SSIM dropping by --1.78\% and PSNR by --1.48 dB, and LPIPS increasing by +0.048. Based on these observations, we adopt $\lambda_{\text{perc}}=1$, $\lambda_{\text{kp}}=5$, and $\sigma=1.0$ as the default hyperparameter configuration in our framework. 
For additional hyperparameter ablations, please refer to Appendix~\ref{appendix:ablation}.

\section{Conclusion}
\label{sec:conclusion}
In this work, we present \textbf{K-Stain}, a framework that addresses intrinsic misalignment in H\&E-to-IHC virtual staining. By leveraging spatial correspondence via keypoints, K-Stain integrates three key modules: the Hierarchical Spatial Keypoint Detector (HSKD), Keypoint-aware Enhancement Generator (KEG), and Keypoint Guided Discriminator (KGD), which together align fine morphological structures while maintaining global staining fidelity. Experiments on two public datasets show that K-Stain surpasses GAN and diffusion-based baselines.

Despite its strong results, keypoint reliability may decline in regions lacking clear landmarks, and the method still depends on paired H\&E–IHC data for training. Future work will explore multimodal extensions, domain robustness via self-supervision or adaptation, and applications to downstream tasks such as biomarker quantification and diagnostic assistance.

\begin{ack}
This work is supported by the Guangdong Science and Technology Department (2024ZDZX2004) and the Guangzhou Industrial Information and Intelligent Key Laboratory Project (No. 2024A03J0628).
\end{ack}

\newpage

\bibliographystyle{unsrtnat}

\appendix

 \section{Additional Ablation on the Number of Keypoints}
\label{appendix:ablation}

We further investigate how the number of detected keypoints $N$ influences the performance of K-Stain. As presented in Table~\ref{tab:keypoint_ablation}, using a small number of keypoints (e.g., $N=32$) provides limited structural anchors, which weakens the spatial guidance and leads to lower SSIM (0.4605) and PSNR (18.92 dB), together with a higher LPIPS (0.2440). Increasing the number of keypoints to $N=64$ improves structural fidelity, yielding moderate gains in both SSIM and PSNR. The best performance is achieved when $N=128$, where the model reaches the highest SSIM (0.4720) and PSNR (19.38 dB), along with the lowest LPIPS (0.2361). However, further increasing $N$ to 256 introduces redundancy and noise in the structural representation, causing a slight performance drop and additional computational cost. These results indicate that a moderate number of keypoints offers the best balance between capturing sufficient structural details and maintaining model efficiency. Therefore, we adopt $N=128$ as the default setting in all experiments.

\begin{table}[h]
\centering
\caption{Ablation study on the number of keypoints $N$.}
\begin{tabular}{c|c c c}
\toprule
$N$ & SSIM $\uparrow$ & PSNR $\uparrow$ & LPIPS $\downarrow$ \\
\midrule
32   & 0.4605 & 18.92 & 0.2440 \\
64   & 0.4689 & 19.21 & 0.2392 \\
128  & \textbf{0.4720} & \textbf{19.38} & \textbf{0.2361} \\
256  & 0.4701 & 19.15 & 0.2375 \\
\bottomrule
\end{tabular}
\label{tab:keypoint_ablation}
\end{table}

\section{Additional Ablation on KSE within KGD}
\label{appendix:appB}
To verify the contribution of the keypoint-to-structural embedding (KSE) used in the Keypoint Guided Discriminator (KGD), we compare KGD with and without the KSE tensor while keeping all other settings identical. The results are summarized in Table~\ref{tab:kse_in_kgd}.

\begin{table}[h]
\centering
\caption{Ablation on the KSE tensor fed into KGD.}
\label{tab:kse_in_kgd}
\vspace{2pt}
\begin{tabular}{lccc}
\toprule
Method & SSIM$\uparrow$ & PSNR$\uparrow$ & LPIPS$\downarrow$ \\
\midrule
KGD w/o KSE & 0.4622 & 18.95 & 0.2418 \\
KGD w/ KSE (ours) & \textbf{0.4720} & \textbf{19.38} & \textbf{0.2361} \\
\bottomrule
\end{tabular}
\end{table}

As shown in Table~\ref{tab:kse_in_kgd}, introducing KSE enables KGD to utilize explicit keypoint information during discrimination, improving structural alignment and demonstrating that keypoints enhance the discriminator’s spatial perception.

\end{document}